\documentclass[journal]{IEEEtran}
\usepackage{booktabs} 
\usepackage{graphicx}
\usepackage{amsmath}

\ifCLASSINFOpdf
\else
\fi
\begin{document}
%
\title{Pseudo-labelling and Meta Reweighting Learning for Image Aesthetic Quality Assessment}
%
%
%

\author{Xin~Jin\thanks{This work is partially supported by the National Natural Science Foundation of China (62072014), the Beijing Natural Science Foundation (L192040), and the CAAI-Huawei MindSpore Open Fund (CAAIXSJLJJ-2021-022A).
		
		Xin Jin is with the Department of Cyber Security, Beijing Electronic Science and Technology Institute, Beijing, China, and also with the Beijing Institute for General Artificial Intelligence (BIGAI), Beijing, China. E-mail: jinxinbesti@foxmail.com.}, 
		Hao~Lou\thanks{Hao Lou, Heng Huang, Xiaokun Zhang and Xiqiao Li are with the Department of Cyber Security, Beijing Electronic Science and Technology Institute, Beijing, China.},
		Huang~Heng, 
		Xiaodong Li*\thanks{Xiaodong Li is with the Department of Cyber Security, Beijing Electronic Science and Technology Institute, Beijing, China. Xiaodong Li* is the corresponding author. E-mail: lxdbesti@163.com.},
		Shuai Cui\thanks{Shuai Cui was with the Department of Philosophy and Mathematics, University of California, Davis, CA, USA.},
		Xiaokun Zhang, 
		Xiqiao Li}
\maketitle

\begin{abstract}
In the tasks of image aesthetic quality evaluation, it is difficult to reach both the high score area and low score area due to the normal distribution of aesthetic datasets. To reduce the error in labeling and solve the problem of normal data distribution, we propose a new aesthetic mixed dataset with classification and regression called AMD-CR, and we train a meta reweighting network to reweight the loss of training data differently. In addition, we provide a training strategy acccording to different stages, based on pseudo labels of the binary classification task, and then we use it for aesthetic training acccording to different stages in classification and regression tasks. In the construction of the network structure, we construct an aesthetic adaptive block (AAB) structure that can adapt to any size of the input images. Besides, we also use the efficient channel attention (ECA) to strengthen the feature extracting ability of each task. The experimental result shows that our method improves 0.1112 compared with the conventional methods in SROCC. The method can also help to find best aesthetic path planning for unmanned aerial vehicles (UAV) and vehicles.
\end{abstract}


%
\IEEEpeerreviewmaketitle

\section{Introduction}
\IEEEPARstart{A}{esthetics} was raised by the German philosopher Bavmgarten in 1750. It focuses on the aesthetic relationship between the public and the world. The aesthetic research subjects are the aesthetic activities, an abstract concept, with strong subjectivity. Only the experts can evaluate the abstractly aesthetic consensus in real life. However, the public can not give the accurately aesthetic evaluation without professional training, which leads to computational aesthetics \cite{datta2006studying}.

The goal of computational aesthetics is aesthetic intelligence that computes and robots are able to recognize, produce, and create beauty. That can make computers and robots focus on human cognition and the inner mechanism of recognizing beauty. Computational visual aesthetics \cite{brachmann2017computational} is an intelligent comprehension about visual information by visual computation techniques. Nowadays, visual computation aesthetics depends on deep learning. In the related researches, the visual computation aesthetics, mainly by training massive data, get neural network modal. So, the model can give the evaluation of aesthetic quality. Although the generally existing aesthetic benchmark dataset like AVA \cite{murray2012ava}, AADB \cite{kong2016photo}, PCCD \cite{chang2017aesthetic} and others have numeric score labels, most of them are not over 20,000 images. AVA is the largest dataset with 25,5530 images, but the image of over 7 and under 3 only take 4.4\% in the overall score. In the training process, that can trigger seriously underfitting problems. AADB includes 9,958 images, and each image have 11 aesthetic attribute scores. However, the aesthetic labels and amount are too few to have good labeling quality. CHUKPQ \cite{luo2011content} only have the simply two-value label instead of the numeric label. The above datasets have obvious disadvantages. Moreover, comparing with target detection, sematic segmentation, semantic classification, and other tasks, it is more challenging to extract aesthetic features in detail. That brings a lot of challenges for evaluation of aesthetic quality. It is hard to extract lighting, color, composition, and other features in aesthetics. That brings serious problems for computational aesthetics.  

In order to solve those problems, we propose a method of aesthetic quality evaluation based on pseudo-labelling and meta-learning. Specifically, we make the following four contributions:

\begin{enumerate}
\item[•] We filtering and constructing an Aesthetic Mixed Dataset with Classification and Regression (AMD-CR) with an appropriate scale. AMD-CR have both more reasonable data distribution and better aesthetic quality labels than the classic aesthetic dataset.

\item[•] We introduce  meta-learning in aesthetic sample re-weight. We use meta reweighting network to provide appropriate weights of the training samples.

\item[•] We propose Aesthetic Adaptive Block (AAB) to compose aesthetic network model that is used to compare images, with ratio between length and width, for effectively evaluation. 

\item[•] To the best of our knowledge, we are the first to use binary classification training pseudo-label in the aesthetic tasks; and, our method shows good results in image aesthetic quality evaluation and path planning for unmanned aerial vehicles (UAV) and vehicles.
\end{enumerate}

The rest of this paper is organized as follows: in Section II we will introduce the related work of image aesthetic quality assessment and meta learning. In Section III we will show the network model and AMD-CR. In Section IV, we will describe the experiments in detail and analyze the results. Finally, Section V will focus on conclusion and future work.

%
%
%
%

\section{Related work}\label{}

\begin{figure*}[btp!]
	\centering
	\includegraphics[width=\linewidth]{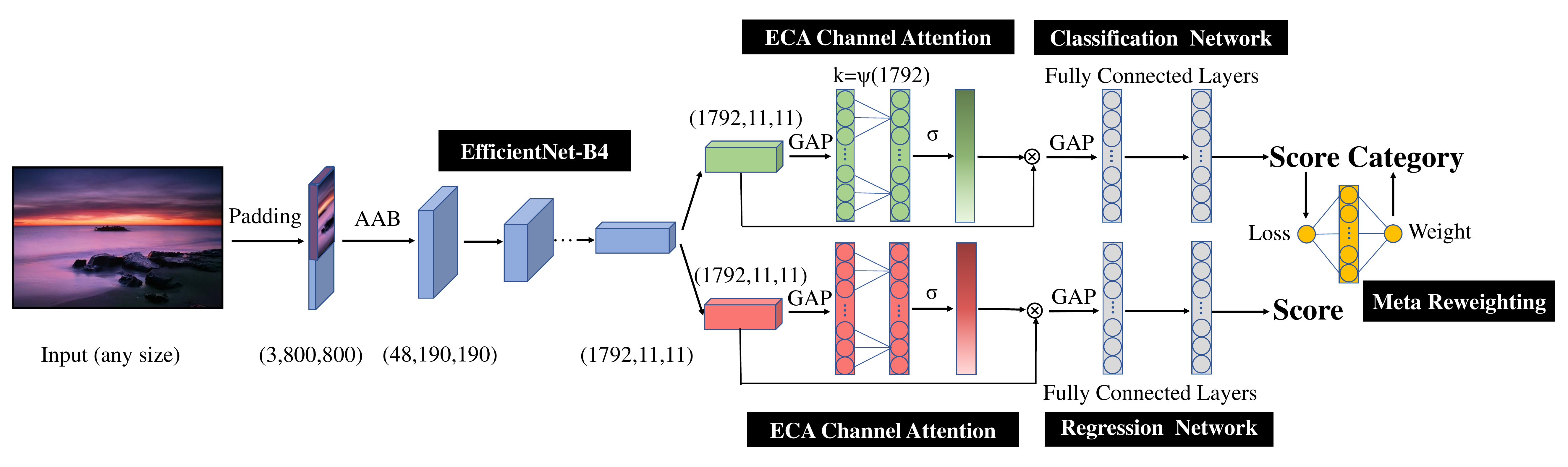}
	\caption{The architecture of the network. The input image size is arbitrary and the main network contains EfficienNet-B4 \cite{tan2019efficientnet}. AAB is the aesthetic adaptive block, GAP is the global average pooling, $\sigma$ is the sigmoid function, meta reweighting network contains a hidden layer of 100 nodes, and each layer will be activated by the relu function.}
	\label{fig:special}  
\end{figure*}

We divide the discussions of related work into the following three subsections.

\subsection{Image aesthetic datasets}
The most common benchmark aesthetic datasets are AVA \cite{murray2012ava} and AADB \cite{kong2016photo}. There are over 250,000 images in AVA. Each image has one numeric aesthetic quality evaluation label, 66 sematic labels, and 14 typed labels. AADB comes from University of California, Irvine. Kong and others designed a new image aesthetic attribute dataset with total 9,958 images. This dataset includes images from both professional photographers and ordinary workers. In fact, there are obvious disadvantages in both datasets. AVA dataset both high score and low score images take less than 5\%. Although AADB includes 11 aesthetic attributes, including balancing elements, colorful harmony, and others, its labels are few and subjective. The limitations of datasets stop the development of image aesthetic quality evaluation. 

\subsection{Image aesthetic quality assessment}
Based on the definition of the classification task in basic aesthetic evaluation tasks, there are two categories of images: high quality and low quality \cite{sperry1969interhemispheric}. However, the judgements are usually based on distinguished aesthetic common sense. If two images are both ordinary images, it will be hard to distinguish them by binary values. So, we need continuously numeric value to evaluate them. In the early stages, researchers designed many handcrafted aesthetics features \cite{tong2004classification, luo2008photo, nishiyama2011aesthetic,marchesotti2011assessing}, Luo et al. \cite{luo2011content} divided 1,7613 images to 7 categories based on semantics; they design a series image detection salient regions method according to different image material. They use machine learning algorithm to train classification to extract features and predict classified results. 

Nowadays, image aesthetic features mainly use deep learning network model \cite{lu2014rapid} to extract automatically. The following researches \cite{kao2015visual,lu2015deep} directly or indirectly model the extra information, such as style; the classification of this model is more accurate than the model based on handcraft. Then, researchers use probability distribution \cite{wu2011learning} to describe aesthetics and explain the subjectivity of aesthetic evaluation. However, the above single image aesthetic vector is hard to quantize complex aesthetic evaluation. Recently, aesthetic attributes evaluation field proposes a method based on hierarchical multi-task network \cite{jin2018predicting}, incremental learning of multi-task network \cite{jin2019incremental}. This method quantize aesthetic evaluation by contrasting composition, light, color, matching, exposure, depth of field, camera use. However, all researches are limited by the amount of data and data distribution. Our research \cite{kuang2019deep} has great improvement in binary classification of aesthetic evaluation and numeric evaluation of regression tasks. Beside the application of basic aesthetic evaluation, we propose a multistream network, composed by spatial, motion, and structural streams, to gain the multimodal features of path planning. That inspires us to explore path planning based on image aesthetic quality evaluation.

\subsection{Meta learning}
In deep learning, the modal has to be retrained by changing scene. Meta learning, learn to learn, can learn the ability of human learning new things by limit data \cite{finn2017model}. Meta learning mainly includes memory \cite{santoro2016meta}, prediction of gradient \cite{andrychowicz2016learning}, attention mechanisms \cite{vinyals2016matching}, LSTM \cite{ravi2016optimization}, reinforcement learning \cite{2003Meta} and others. By the inspiration of meta learning \cite{ravi2016optimization,finn2017model}, the recent researches study reweight cases adaptatively from data sample. That shows learning is more automatic and dependent. This idea can effectively solve the longtail problem caused by few data taggers.

\section{Resarch methodology}\label{}

In this section, we mainly introduce the structure and the algorithm of the aesthetic assessment model in detail. The network structure is constructed in Fig.1.

\subsection{Meta reweighting network}
\label{sec:3.1}
The loss functions of conventional deep learning are generally cross entropy or mean square error (MSE) loss function, which calculate the loss of average on batch size samples; this method is easily influenced by the sample distribution. In the classificational task, if the proportion of the positive samples is high in samples, then the final transmission of the loss value must be influenced by most of the positive sample loss. Although a few negative samples have higher loss, the network will be less influenced by these negative samples. The idea of the meta reweighting in this paper is to learn the loss weight of the data automatically,  optimize the normal data distribution, and enhance the model robustness.

By the meta reweighting method mentioned in \cite{shu2019meta}, we apply this method to the aesthetic quality assessment tasks. The idea of meta reweighting is to minimize the loss $\frac{1}{N}\sum_{i=1}^{N}\ell\left(y_i,{f}(x_i)\right)$ in the training set and extract the best classifier parameters $w^{*}$. In the biased training set, each sample weight is expressed as $\mathcal{V}\left(L_i^{train}(w);\mathrm{\Theta}\right)$,  $L_i^{train}\left(w\right)=\ \ell\left(y_i,{f}(x_i)\right)$, and it means the loss of \emph{i}th training sample; \emph{$\Theta$} represents the parameter of the meta reweighting network(MRN). The best parameter $w^{*}$ is calculated as Equation (1).
\begin{equation}
	w^\ast(\Theta)=\frac{1}{N} \sum_{i=1}^{N}\mathcal{V}\left(L_{i}^{\mathrm{train}}\left(w\right);\Theta\right)L_{i}^{\mathrm{train\ }}\left(w\right)
\end{equation}

In order to learn hyperparametric $\Theta$ automatically, we generate a multilayer perceptron based on the hidden layer in the method of meta learning, shown as the yellow network in Fig.1. The input of the meta reweighting network is the loss of samples, and the output is the weight of the loss of samples. We extract a few high quality images from the training set \emph{S$_{train}$} to construct the meta set \emph{S$_{meta}$}. \emph{N} is the amount of \emph{S$_{train}$}, and \emph{M} is the amount of \emph{S$_{meta}$}, \emph{N} $\gg$ \emph{M}. The advantages of \emph{S$_{meta}$} are correct labels and equal data distribution. According to the score labels of aesthetic quality assessment datasets, we divide the data into 10 segments equally; we extract 200 images from each segment artificially. If the amount of any segment is less than 200, we will choose appropriate amount of  images from the adjacent segments to replenish the lack images. Finally, we filter 2000 images fo construct \emph{S$_{meta}$}. There are two stages in the training process; we update the \emph{$\Theta$} and $w$ through a single optimization cycle. We firstly update the parameter \emph{$\Theta$} of the MRN, and then update the parameter $w$ of the main network in each iteration optimization. The update equation for the first stage is defined as Equation (2), \emph{$\beta$} is the step size of the MRN.

\begin{equation}
	\Theta^{(t+1)}=\Theta^{(t)}-\left.\beta\frac{1}{m}\sum_{i=1}^{m}\nabla_\Theta L_i^{meta}\left({\hat{w}}^{(t)}(\Theta)\right)\right|_{\theta^{(t)}}
\end{equation}

$\left.\nabla_\Theta L_i^{meta}\left({\hat{w}}^{(t)}(\Theta)\right)\right|_{\theta^{(t)}}$ represents the gradient of the MRN on meta set, ${\hat{w}}^{(t)}\left(\theta\right)$ is calculated as Equation (3), \emph{$\alpha$} is the step size of the main network.

\begin{equation}
	\begin{aligned}
		&{\hat{w}}^{\left(t\right)}\left(\theta\right)= \\
		&w^{\left(t\right)}-\left.\alpha\frac{1}{n}\sum_{i=1}^{n}v\left(L_i^{train}\left(w^{\left(t\right)}\right);\Theta\right)\nabla_\mathbf{w}L_i^{\mathrm{train}}\left(w\right)\right|_{w^{\left(t\right)}}
	\end{aligned}
\end{equation}

In the first stage, we need to calculate the weighted loss of \emph{S$_{train}$} on the main network. In order to distinguish real training from meta-learning training, we copy the parameters of the main network at each single cycle training. The parameters of MRN are optimized by Equation (2).

In the second stage, we optimize and update the main network according to the loss of \emph{S$_{train}$} and the weight from MRN on the meta network, it is as Equation (4).
\begin{equation}
	\begin{aligned}
		&w^{(t+1)}=w^{(t)}- \\
		&\left.\alpha\frac{1}{n}\sum_{i=1}^{n}\nu\left(L_i^{train}\left(w^{\left(t\right)}\right);\Theta^{\left(t+1\right)}\right)\nabla_\mathbf{w}L_i^{train}\left(w\right)\right|_{w^{\left(t\right)}}
	\end{aligned}
\end{equation}

\subsection{Aesthetic adaptive block}
\label{sec:3.2}
\begin{figure}[btp!]
	\centering
	\includegraphics[width=\columnwidth]{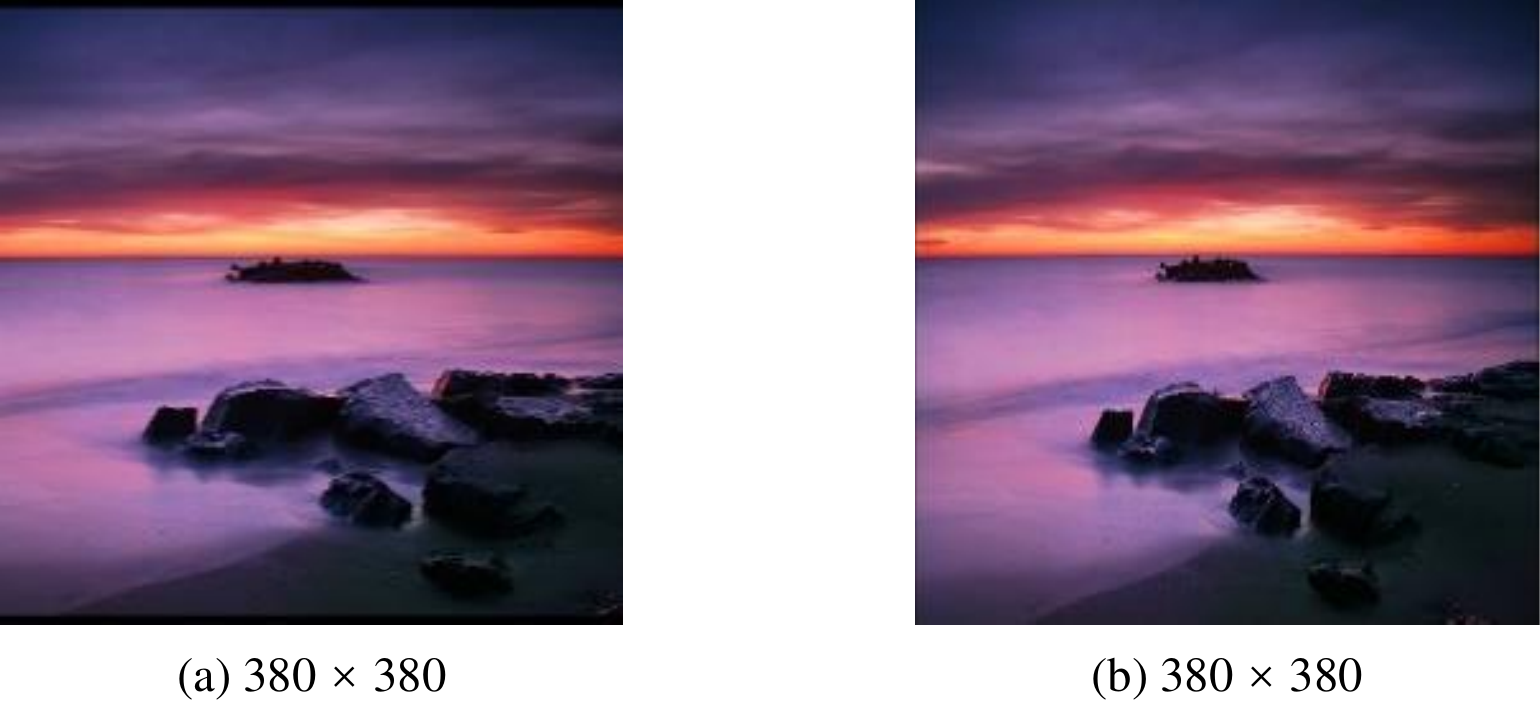}
	\caption{(a) is the image for cropping, and (b) is for resizing.}
	\label{fig:2}  
\end{figure}

\begin{figure}[btp!]
	\centering
	\includegraphics[width=\columnwidth]{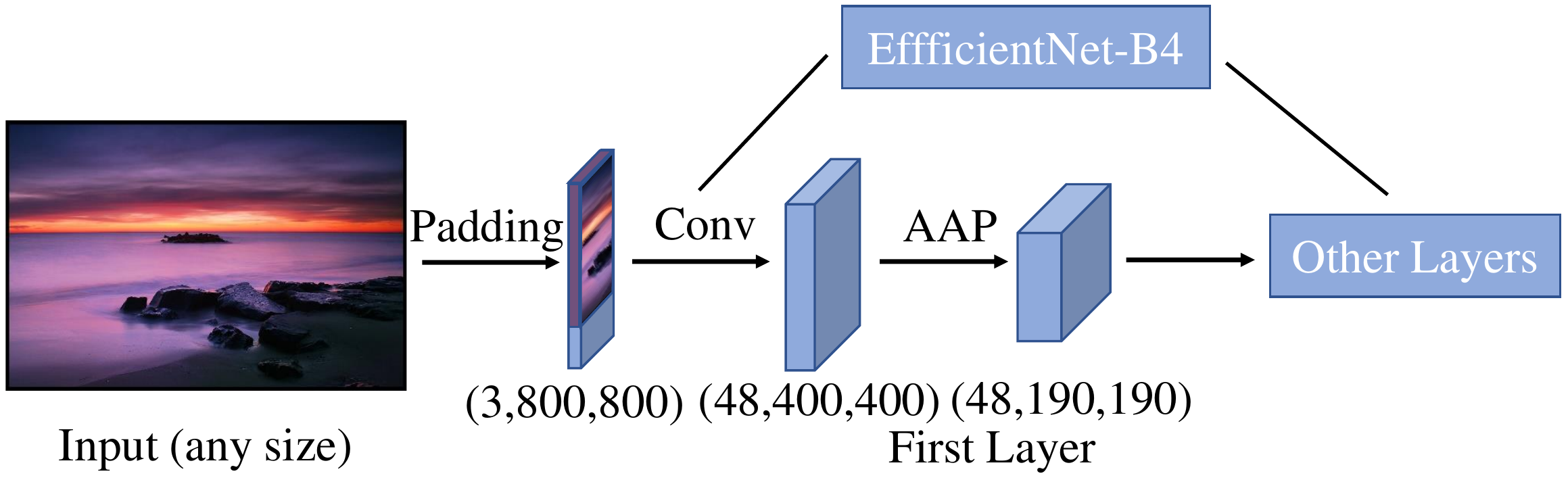}
	\caption{The structure of the aesthetic adaptive block, AAP is the adaptive average pooling layer.}
	\label{fig:3}  
\end{figure}

For the pretraining of images, there are three ways: cropping, resizing and adaptive padding. The best input size of images in EfficientNet-B4 is 380×380. For cropping, we adjust any size images to make the short edge reach 380 proportionally, and then crop the center of images into the best input size. For resizing, we stretch any size images to 380×380 instead of cropping. Both methods may lose a part of the aesthetic features. We propose a structure, aesthetic adaptive block (AAB), based on adaptive padding; this method can adapt any input size without cropping and resizing. In AAB, we adjust the long edge of the image to 800 proportinally, and then we fill the short edge with 0 to make it reach 800. So, we construct an 800x800 input image. Then, for the first convolution operation of the EfficientNet-B4, we use an adaptive average pooling layer to reshape the feature map of 48×190×190; 48 is the amount of the feature channel. Finally, the feature map enters the remaining network layers in EfficientNet-B4. Three methods are shown as Fig.2 and Fig.3.

\subsection{Eifficent channel attention}
\label{sec:3.3}
Recently, the channel attention mechanism \cite{hu2018squeeze} shows a great potential in improving the performance of deep learning network models. However, most of the previous focused on building complex attention modules to achieve good performance. In order to reduce the complexcity of models, this paper cites the ECA module \cite{wang2020eca}. It adopts a local cross-channel attention mechanism and the adaptive one-dimensional convolution kernel, which can improve performance obviously.

The ECA captures local cross-channel attention information by interacting with each channel and its \emph{k} neighbors. This can be effectively implemented by a \emph{1D} convolution with the kernel size of \emph{k}, where \emph{k} represents the local cross-channel attention and means the number of neighbor nodes involved in the attention prediction of this channel. In order to avoid adjusting \emph{k} in the validation phase, the ECA proposes a method that
 \emph{k} can be adaptively proportional to the channel dimensions, \emph{k} is expressed as Equation (5).
\begin{equation}
	k=\psi(C)\ =\left|\frac{\log_2(C)}{\gamma}+\frac{b}{\gamma}\right|_{\mathrm{odd}}	
\end{equation}

${|t|}_{odd}$ means the nearest odd number of \emph{t}, the value of \emph{$\gamma$} is 2, the value of \emph{b} is 1. We can get \emph{k} = 7 if \emph{C} = 1792. 

\subsection{Multi-task network}
\label{sec:3.4}
Both the aesthetic multi-classification and regression tasks share the same main network parameters to extract features. The multi-classification network and the regression network structure are similar as shown in Fig.1. The main network outputs the \emph{N}×1792×11×11 feature map. \emph{N} is for the batch size. The number of channels is first reduced to 448 using a convolution layer with the 3×3 kernel, then the feature map is transformed into fully connected layer in 448 dimensions by an adaptive average pooling layer. Finally, the last layer has 10 nodes in the multi-classification network and 1 node in the regression network. Loss function formula of multi-classification network is expressed as Equation (6).
\begin{equation}
	{Loss}_{class}=\sum_{x}p(x)logq(x)	
\end{equation}

In this equation, \emph{p(x)} represents the predicted output probability, \emph{q(x)} represents the ground-truth probability, and the probability is obtained by the following Equation (7).
\begin{equation}
	y_i=\frac{e^i}{\mathrm{\Sigma}_je^j}	
\end{equation}

Loss function formula of regression network is expressed as Equation (8).
\begin{equation}
	Loss_{regre}=\frac{1}{N}\sum_{i}{(s_i-\widehat{s_i})}^2	
\end{equation}

\emph{$s_{i}$} is the output score, \emph{$\hat{s_{i}}$} is the ground-truth score.

\subsection{Piecewise strategy based on pseudo-labelling of the binary classification task}
\label{sec:3.5}
The pseudo-labelling was firstly proposed in the \cite{hinton2015distilling}, which are not completely consistent with the ground-truth labels; and. the ground-truth labels are often used as the hard labels. Pseudo labels can be considered as soft labels after training. We propose a piecewise training method based on pseudo labels of the binary classification task. For first stage, we train a binary classifier, and the images are divided into \emph{0} and \emph{1} based on the boundary of 5 points, while 0 represents low aesthetic quality and 1 represents high aesthetic quality. TABLE I shows the accuracy and error rates of each segment in the testing set.

\begin{table}[htbp]
	\caption{Correctness and error rates for classification}
	\centering
	\label{tab:1}
	\begin{tabular}{ccc}
		\hline\noalign{\smallskip}
		Score Segment & Correctness rate & Error Rate  \\
		\noalign{\smallskip}\hline\noalign{\smallskip}
		0.0-1.0 & 100\% & 0\%  \\
		1.0-2.0 & 100\% & 0\%  \\
		2.0-3.0 & 98.67\% & 1.33\%  \\
		3.0-4.0 & 92.85\% & 7.15\%  \\
		4.0-5.0 & 70.06\% & 29.94\%  \\
		5.0-6.0 & 75.73\% & 24.27\%  \\
		6.0-7.0 & 91.42\% & 8.58\%  \\
		7.0-8.0 & 96.01\% & 3.99\% \\
		8.0-9.0 & 97.01\% & 2.99\% \\
		9.0-10.0 & 100\% & 0\% \\
		\noalign{\smallskip}\hline
	\end{tabular}
\end{table}

Although there are still a few false classification samples in  2.0-3.0 and 8.0-9.0 intervals, the error rate in the middle interval 4.0-6.0 is higher; it is because, in the middle interval, there are no huge difference in the scores of images in all perspectives of aesthetic quality assessment. However, this does not influence the final assessment of image aesthetics by the network model. Through continuous score regression, the assessment measures of the model can be measured by quantitative errors. If the scores of data is subjective labels, the result of the classification task by the binary classifier can be regarded as pseudo labels of the model. In the second stage, the pseudo-labelling redivides the datasets into two datasets. We performed multi-classification training and fine-grained regression training separately to extract better image aesthetic features. The experimental results in Section IV show that the piecewise strategy helps to improve the accuracy of the regression.

\subsection{Data}

In order to construct an aesthetic dataset with reasonable distribution and high quality of aesthetic annotation, we filter and reconstruct a dataset called the aesthetic mixed dataset with classification and regression (AMD-CR). There are two datasets in AMD-CR: aesthetic mixed dataset with classification (AMD-C) applied to the binary classification training task, and aesthetic mixed dataset with regression (AMD-R) applied to the regression task.

\subsubsection{Construction of AMD-CR}

The part of aesthetic mixed dataset with classification and regression (AMD-CR) comes from image aesthetic benchmark datasets: DPChallenge, Photo.net, AVA \cite{murray2012ava},  CUHKPQ \cite{luo2011content} and SPAQ \cite{fang2020perceptual}; another part comes from self-built datasets: GLAMOUR, OUTDOOR, NG and PSA. AVA and DPChallenge come from www.dpchallenge.com. So,  we need to delete the duplicate images. Photo.net is from www.photo.net, GLAMOUR is from www.glamour-photos.org, OUTDOOR is from www.outdoor-photos.com, NG is from www.dili360.com, PSA is from www.psachina.org. We obtain a portion of the public images from these websites and integrate them into our aesthetic dataset. We distribute continuous labels to [0, 10.0] through standardized fractional labeling and map binary labels to \emph{0} and \emph{1}. \emph{0} represents the images with low aesthetic quality and \emph{1} represents the images with high aesthetic quality. Discrete datasets only have binary labels, while continuous datasets both have binary labels and contiuous labels. We focused on selecting images of high and low quality, and we performed a preliminary selection of the images with general aesthetic quality to balance the data distribution. TABLE II shows the constitution of the aesthetic mixed dataset with classification and regression (AMD-CR). 

\begin{table}[htbp]
	\caption{The constitution of AMD-CR}
	\centering
	\label{tab:2}
	\begin{tabular}{ccc}
		\hline\noalign{\smallskip}
		Datasets & Volume & labels  \\
		\noalign{\smallskip}\hline\noalign{\smallskip}
		DPChallenge + AVA & 61461 & Continuous  \\
		CUHKPQ & 29690 & Discrete  \\
		Photo.net & 8985 & Continuous  \\
		NG & 4140 & Discrete  \\
		SPAQ & 4102 & Continuous  \\
		PSA & 3460 & Discrete  \\
		GLAMOUR & 2701 & Discrete \\
		OUTDOOR & 1916 & Discrete \\
		\noalign{\smallskip}\hline
	\end{tabular}
\end{table}

\subsubsection{Construction of AMD-R}

We fliter the data with the continuous labels in the aesthetic mixed dataset with regression (AMD-CR). The images of low and high segments with the scores less than 4.0 points or more than 6.0 points are reserved, and the middle segment images (4.0-6.0) are randomly sampled. Then we get 59,371 images form AMD-R. The segment distribution of AMD-R is shown in Fig.4.
\begin{figure}[btp!]
	\centering
	\includegraphics[width=\columnwidth]{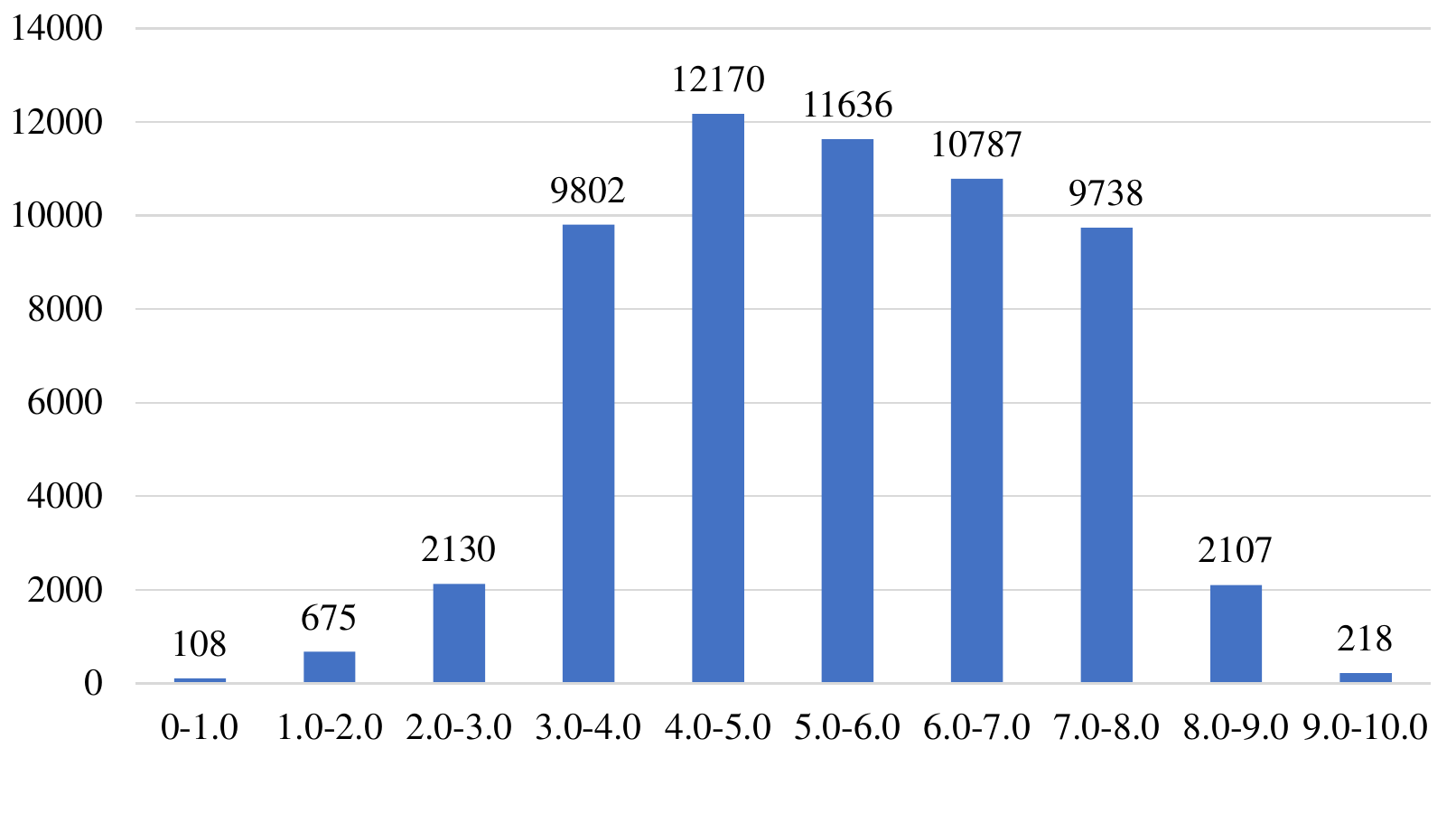}
	\caption{The Segment Distribution Of AMD-R}
	\label{fig:4}  
\end{figure}
\subsubsection{Construction of AMD-C}

We removed the images ranging between 4.0 points and 6.0 points in the AMD-CR to ensure that the ratio of positive and negative samples is 1:1. Finally, we obtain the aesthetic mixed dataset with classification (AMD-C) containing 61,660 images.

\section{Experiment}\label{}

\subsection{Training details}
We set the classification batch size to be 32, the regression batch size to be 64 and the learning rate to be 0.0001. We use Adam as the optimizer; betas are set as (0.98, 0.999); weight decay is set as 0.0001. If either the accuracy rate of classification is not improved or regression loss is not decreased in two consecutive rounds in validation datasets, the learning rate will multiply by 0.5. Our running environment is in Pytorch 1.5.0 and Nvidia TITAN XPs.

The datasets used in this paper are AMD-C and AMD-R. We divide the dataset into three sets; the ratio of training set and validation set and testing set is 8:1:1. EfficientNet-B4 is the main network in the model. After we first trained the binary classifier, we will get the training model as \emph{C$_{2}$}. Then the training set and validation set are divided into \emph{S$_{train0}$}, \emph{S$_{train1}$}, \emph{S$_{valid0}$}, and \emph{S$_{valid1}$} according to the classifier, \emph{0} represents the low quality and \emph{1} represents the high quality in aesthetics. Then we use \emph{S$_{train0}$}, \emph{S$_{valid0}$} and \emph{S$_{train1}$}, \emph{S$_{valid1}$} to train classification and regression models for the next stage.

In the aesthetic assessment training. We start with a 10-class classification training, with [0,10.0] divided into ten average segments, corresponding to the category \emph{A} when the scoring label is in the (\emph{A}, \emph{A}+1.0] interval, \emph{A} $\in$ (0,1,2,3,4,5,6,7,8,9). If there is an image with 0 score, it is classified as the \emph{0} category. When we perform the ten-class trainning task with the images of the same pseudo labels obtained by the binary classifier, the parameters of the regression network should not be returned. We use cross-entropy function to get classification loss. Then, we release the regression network and freeze the main network, classification network, and MRN to train the regression model, we can get models \emph{R$_{0}$} and \emph{R$_{1}$} corresponding to the \emph{S$_{train0}$} and \emph{S$_{train1}$} respectively. In the same method, we can get regression model \emph{R$_{all}$} training on all the AMD-R. Finally we can get the result score like this: if $C_2(i)$ = 0,  $score=\frac{R_0(i)+R_{all}(i)}{2}$,else $score=\frac{R_1(i)+R_{all}(i)}{2}, i\in S_{test}$.

\begin{figure*}[btp!]
	\centering
	\includegraphics[width=\linewidth]{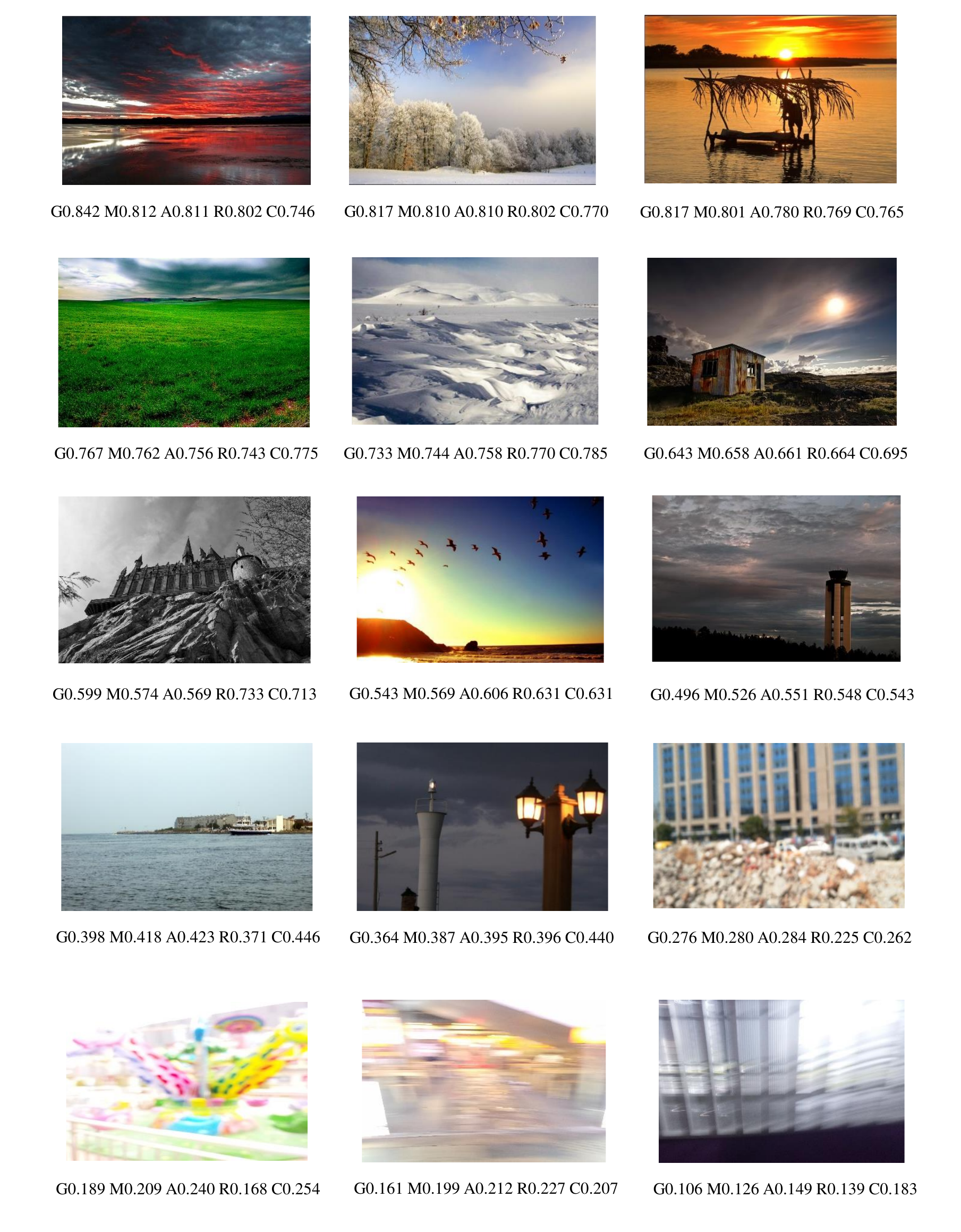}
	\caption{Test samples. \emph{G} for the ground-truth score,  \emph{M} for the score of MRN+AAB+ECA(PCR), \emph{A} for the score of AAB+ECA(PCR), \emph{R} for the score of Resizing+ECA(CR) and \emph{C} for the score of Cropping+ECA(CR).}
	\label{fig:5}  
\end{figure*}
\begin{figure*}[htbp]
	\centering
	\includegraphics[width=\linewidth]{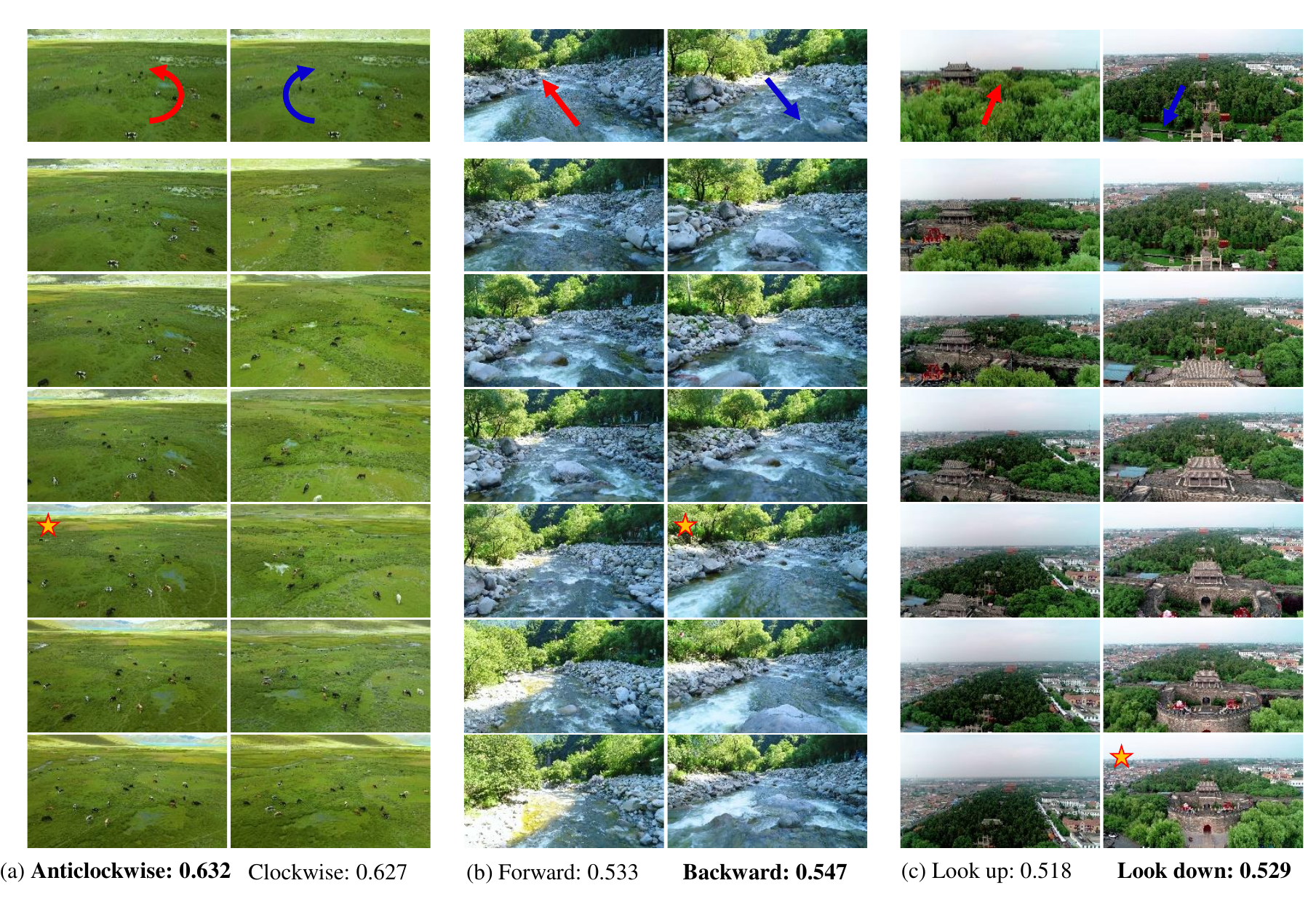}
	\caption{Path planning samples. The better path is identified in bold and the best capture in this path is identified by a yellow star. }
	\label{fig:6}  
\end{figure*}

\subsection{Analysis of experimental results}

We use these indicators for assessment: first, we calculate mean square error (MSE): calculating the loss between the output score and the ground-truth score, it is as Equation (9):
\begin{equation}
	MSE=\frac{1}{N}\sum_{i}{(r_i-\widehat{r_i})}^2
\end{equation}

Mean absolute error (MAE) measures the differences between ground-truth and regressed scores directly, and then find the mean of differences, it is as  Equation (10):
\begin{equation}
	MAE=\frac{1}{N} \sum_{i}\left|r_{i}-\hat{r}_{i}\right|
\end{equation}

We caluculate Spearman’s rank order correlation coefficient (SROCC): the correlation between the output score and the ground-truth score, it is as Equation (11):
\begin{equation}
	SROCC=1-\frac{6\sum_{i}{(r_i-\widehat{r_i})}^2}{N^3-N}
\end{equation}

Accuracy indicates whether the prediction score is consistent with the ground-truth score in the binary classification task when the dividing line is \emph{5}, it is as Equation (12):
\begin{equation}
	ACCURACY=\frac{TP+TN}{P+N}
\end{equation}

$ACCURACY_{\left|error\right|\le1}$ indicates whether the absolute value of the error between the output score and the ground-truth score is within 1 point, it can be expressed as Equation (13):
\begin{equation}
	ACCURACY_{\left|error\right|\le1}=\frac{N_{\left|error\right|\le1}}{N}
\end{equation}

Several experiments were conducted as follows: for training strategies, we record \emph{R} for the single regression training method, \emph{CR} for the classification before regression training method, and \emph{PCR} for classification before regression based on pseudo-labelling training in TABLE III. We verify the goodness of guidance for classification to regression  firstly. We use the cropping method in Section III; we can obviously compare \emph{Cropping(R)} with \emph{Cropping(CR)} to find that if classification before regression will get better result on the testing set. Then we verify that the ECA plays a great role in the aesthetic assessment task. The comparison experiments of \emph{Cropping(CR)} and \emph{Cropping+ECA(CR)} can verify that the ECA can effectively improve the accuracy of the assessment. Besides, we can verify the training method of pseudo labels can improve the result by \emph{AAB+ECA(PCR)}. Finally, we can also verify the meta reweighting network can improve the result by \emph{MRN+AAB+ECA(PCR)}, all the ablation experiments results are shown in Table III. 

\begin{table}[htbp]
	\caption{Ablation experimental results}
	\centering
	\label{tab:4}
	\begin{tabular}{cp{0.7cm}cp{0.8cm}c}
		\hline\noalign{\smallskip}
		Method & MSE↓ & SROCC↑ & Acc↑ & $Acc_{\left|error\right|\le1}$↑  \\
		\noalign{\smallskip}\hline\noalign{\smallskip}
		\textbf{MRN+AAB+ECA(PCR)} & \textbf{0.6691} & \textbf{0.8471} & \textbf{86.53\%} & \textbf{80.61\%}  \\
		AAB+ECA(PCR) & 0.6888 & 0.8409 & 86.48\% & 79.80\%  \\
		AAB+ECA(CR) & 0.7575 & 0.8238 & 84.58\% & 78.73\%  \\
		Resizing+ECA(CR) & 0.7684 & 0.8213 & 84.82\% & 78.23\%  \\
		Cropping+ECA(CR) & 0.8213 & 0.8117 & 84.43\% & 76.54\%  \\
		Cropping(CR) & 1.0325 & 0.7743 & 79.23\% & 69.58\%  \\
		Cropping(R) & 1.2520 & 0.7359 & 76.39\% & 63.78\%  \\
		\noalign{\smallskip}\hline
	\end{tabular}
\end{table}
\begin{table}[htbp]
	\caption{Comparisons on AADB dataset}
	\centering
	\label{tab:4}
	\begin{tabular}{cccc}
		\hline\noalign{\smallskip}
		Method & MAE↓ & SROCC↑ & Acc↑ \\
		\noalign{\smallskip}\hline\noalign{\smallskip}
		Reg-Net \cite{kong2016photo} & 0.1268 & 0.678 & - \\
		RGNet \cite{liu2020composition} & - & 0.710 & - \\
		Lee \emph{et al.} \cite{lee2019image} & 0.1141 & \textbf{0.879} & - \\
		\hline\noalign{\smallskip}
		Ours & \textbf{0.1016} & 0.7185 & \textbf{80.5\%}  \\
		\hline\noalign{\smallskip}
	\end{tabular}
\end{table}

\begin{figure}[t]
	\centering
	\includegraphics[width=\linewidth]{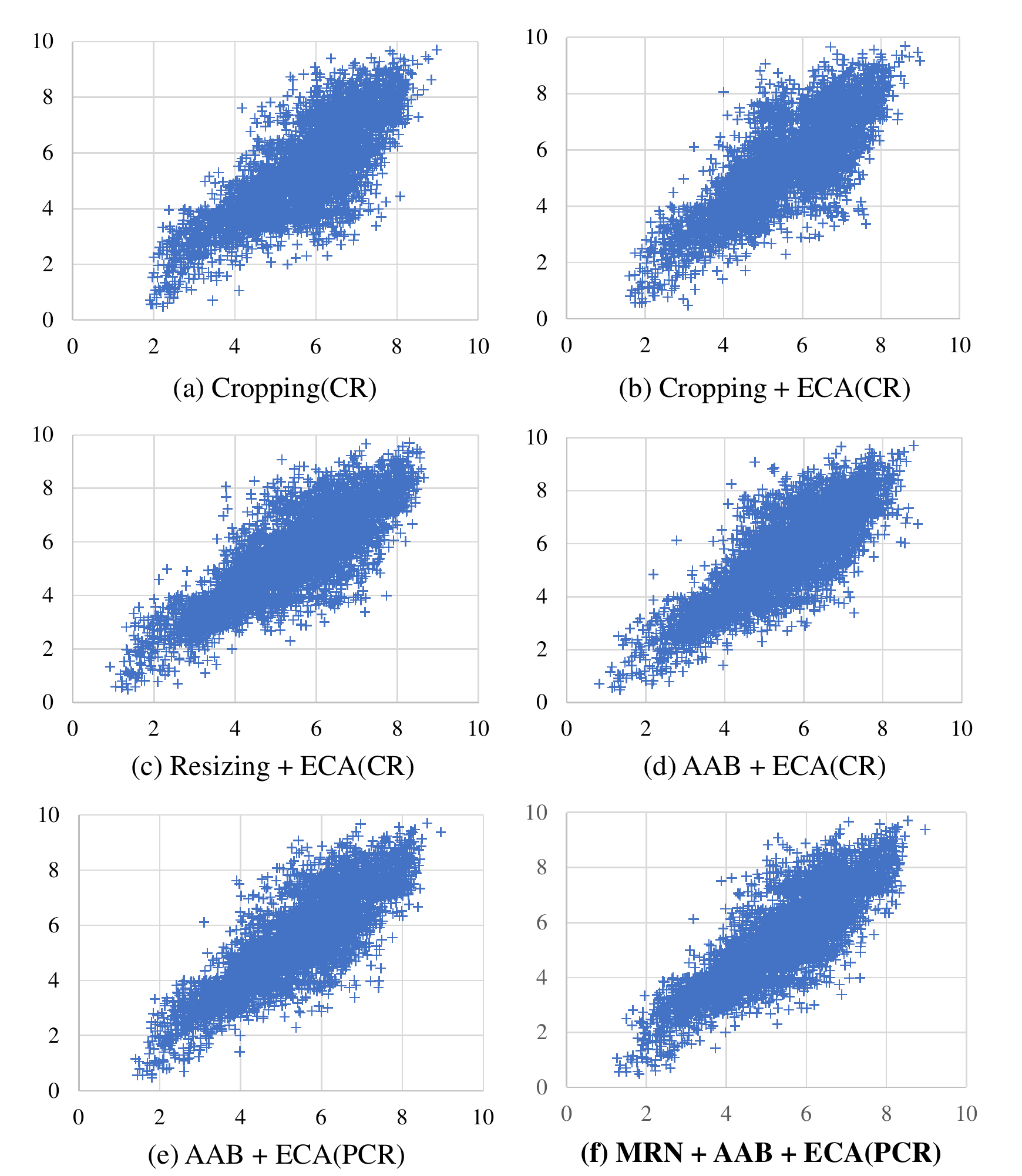}
	\caption{Comparison of scatter plots. 6(a) represents Cropping(CR), 6(b) represents Cropping + ECA(CR), 6(c) represents Resizing + ECA(CR), 6(d) represents AAB + ECA(CR), 6(e) represents AAB + ECA(PCR), and 6(f) represents MRN + AAB + ECA(PCR). The horizontal axis represents the prediction scores, and the vertical axis represents the ground-truth scores. }
	\label{fig:7}  
\end{figure}

Fig.5 compares the test score results for four main methods. We select some typical samples in each score segment: \emph{G} represents the ground truth; \emph{M} represents the score of \emph{MAE(PCR)}; \emph{A} represents the score of \emph{AE(PCR)}; \emph{R} represents the score of \emph{RE(CR)}; \emph{C} represents the score of \emph{CE(CR)}. It can be found that each segment of the dataset has obvious distinction, and \emph{MAE(PCR)} method has the best ability for asssessment. In TABLE IV, by comparison, our method has better performance in MAE and accuracy than others.

Fig.6 shows the path planning samples; we choose three different scenarios, and each scenary has two different path planning strategies. It is easy for our model to select one of the path with high aesthetic quality assessment. Besides, we choose the image with the highest aesthetic score for each selected path. These images are all better in color, light, composition and other aesthetic photography features.

We can get the prediction of the distribution with different methods from the scatter plots of Fig.7. Fig.7 shows that the best model can predict the scores between 1 and 9 points, which is very close to the actual label score interval; the prediction results of the high and low segments are ideal; (f) shows a long and straight spinning cone, that means (f) has the best distribution for aesthetic regression.

\begin{figure}[htbp]
	\centering
	\includegraphics[width=\columnwidth]{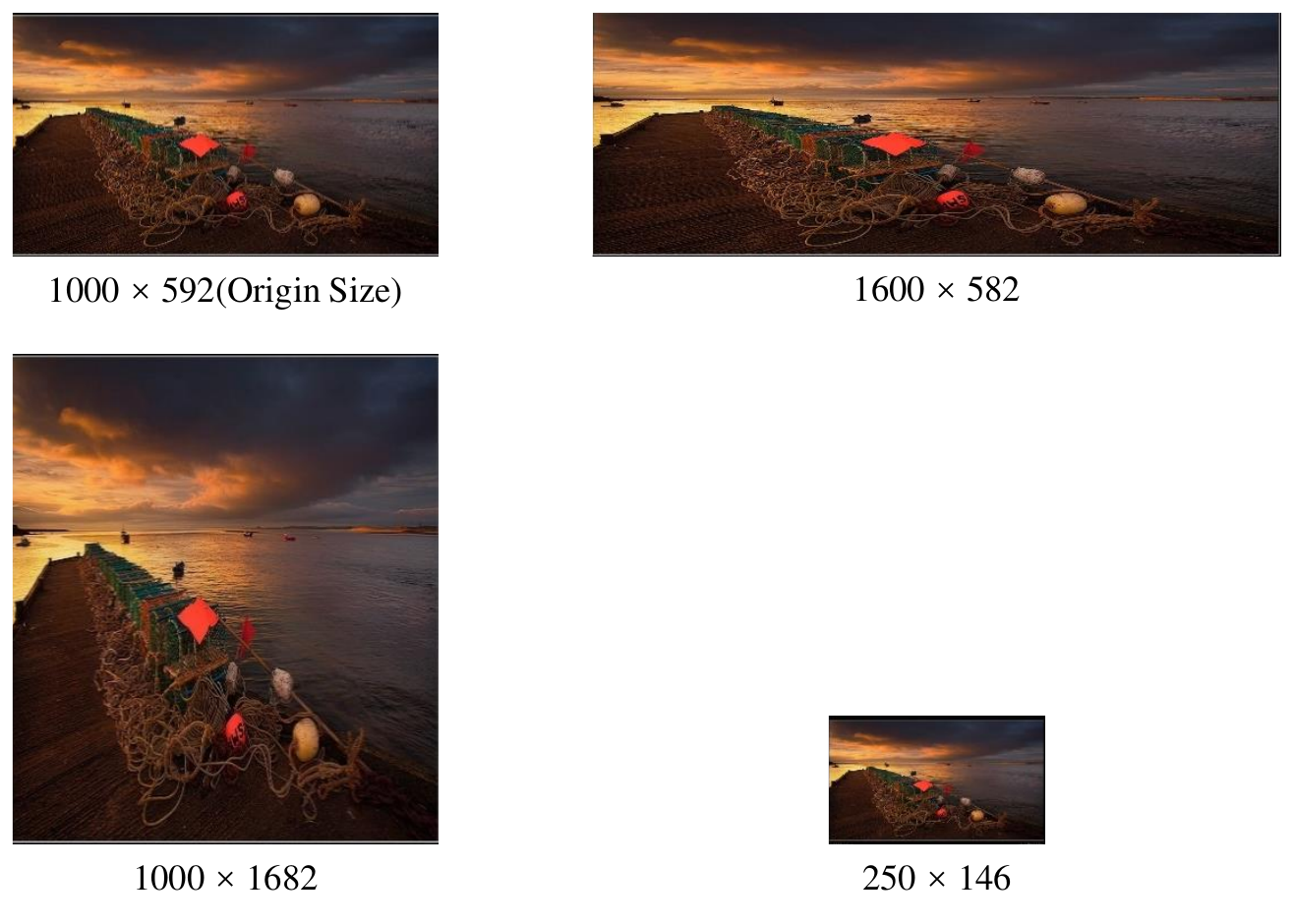}
	\caption{Images scaling in different sizes}
	\label{fig:8}  
\end{figure}

\begin{table}[htbp]
	\caption{Prediction results of different image preprocessing methods}
	\centering
	\label{tab:3}
	\begin{tabular}{ccccc}
		\hline\noalign{\smallskip}
		Image size & Origin & AAB & Resizing & Cropping  \\
		\noalign{\smallskip}\hline\noalign{\smallskip}
		1000×582 & 8.12 & 8.14 & 8.21 & 6.78  \\
		1600×582 & 8.12 & 7.67 & 8.21 & 6.97  \\
		1000×1782 & 8.12 & 7.80 & 8.29 & 7.33  \\
		250×146 & 8.12 & 7.83 & 8.02 & 7.23 \\
		\noalign{\smallskip}\hline
	\end{tabular}
\end{table}

Comparing with three different image preprocessing methods in Section III.B, we can find that AAB method can save the information about the length and the width in the original image. Resizing can save the whole picture information, but, when the ratios between length and width are either extremly high or extremely low, resizing may lead to the loss of aesthetic features. Cropping only cuts the center into 380×380, which may lose some important image feature information. As shown in the Fig.8.

Table V shows that the original image is stretched horizontally to 1600×592, longitudinally to 1000×1792, and isometric scaling to 250×146. AAB model has different output in different ratios of length and width. The scores of resizing are closer, and the cropping model has low scores. It is because both methods may neglect of local features.

\section{Conclusions}\label{}

It is a challenge task to construct a new dataset in image aesthetic field. We construct AMD-CR with reasonable classification regression by mixing and filtering massive datasets; we proposed a new aesthetic evaluation model based on meta reweight network and binary classification pseudo labels. This model can effectively improve ability of predict image aesthetic. Moreover, we design AAB structure to use in any input size images. We are also the first team use ECA in aesthetics. This attention can improve ability of extracting features without changing the number of channels. The best test set, SROCC, is 0.8471 that is higher than classic deep learning regression model.

Now, we only evaluate images generally/ In the future, we will improve application aesthetic quality evaluation, Interpretability, by more image attributes. We also consider how to use emotions, arts, themes, and other high semantic information to help aesthetic evaluation tasks. Moreover, we hope to explore more in the area of aesthetic guidance.

\ifCLASSOPTIONcaptionsoff
  \newpage
\fi


\bibliographystyle{IEEEtran}
\bibliography{ReferenceData.bib}

%
%
%

%

\begin{IEEEbiography}[{\includegraphics[width=1in,height=1.25in,clip,keepaspectratio]{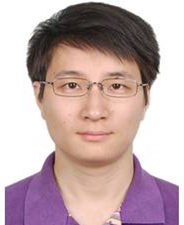}}]{XIN JIN}
was born in Anhui, China. He received the Ph.D. degree from Beihang University, China. He is currently an Associate Professor with the Department of Cyber Security at Beijing Electronic and Science Technology Institute and a Visiting Scholar at Beijing Institute for General Artificial Intelligence (BigAI). He received his Ph.D. degrees in Computer Science from Beihang University, China, in 2013. He was a visiting student at Lotus Hill Institute, China and a visiting scholar at Tsinghua University, Beijing China. His research interests include computational aesthetics, computer vision and artificial intelligence security. He served as the program committee members and session chairs of multiple conferences including AAAI 2020-2022, IJCAI 2020, ACM MM 2021, ISAIR 2017-2022, ICVRD2017-2020. He has received the Best Paper Awards at ROSENET 2018/2019.
\end{IEEEbiography}

\begin{IEEEbiography}[{\includegraphics[width=1in,height=1.25in,clip,keepaspectratio]{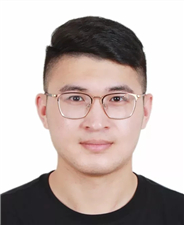}}]{HAO LOU}
	was born in Zhejiang, China. He received the bachelor's degree from Northwest Agriculture and Forestry University, China. He is currently pursuing the master’s degree with the Beijing Electronic Science and Technology Institute, China. His research interest includes visual media aesthetics.
\end{IEEEbiography}

\begin{IEEEbiography}[{\includegraphics[width=1in,height=1.25in,clip,keepaspectratio]{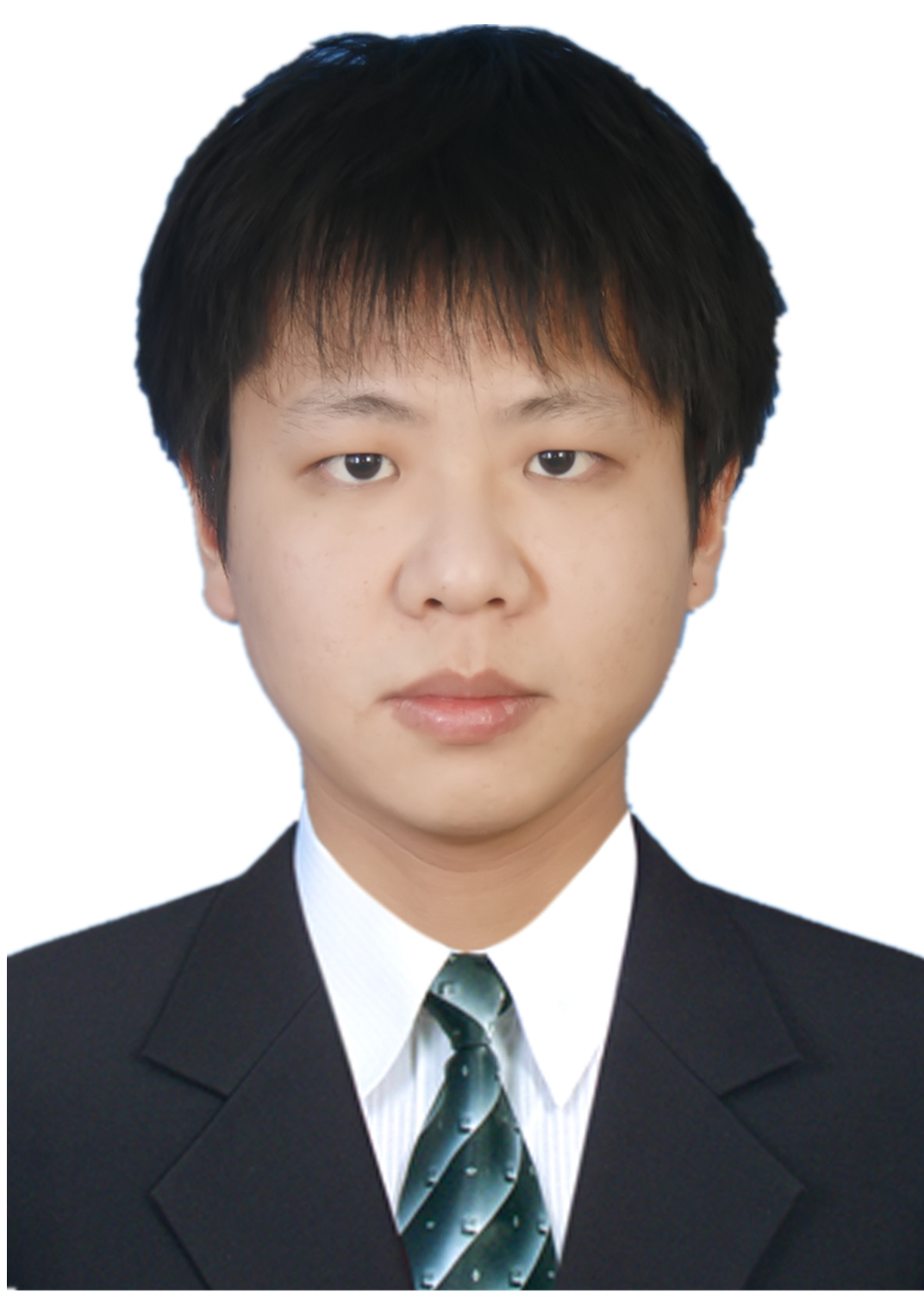}}]{HUANG HENG}
	was born in Fujian, China. He received the bachelor's degree from Shangdong University, China. He is currently pursuing the master’s degree with the Beijing Electronic Science and Technology Institute, China. His research interest includes visual media aesthetics.
\end{IEEEbiography}

\begin{IEEEbiography}[{\includegraphics[width=1in,height=1.25in,clip,keepaspectratio]{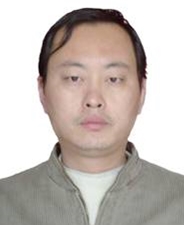}}]{XIAODONG LI}
	was born in Henan, China. He received the Ph.D. degree from Northwestern Polytechnic University, China. He is currently an Associate Professor with the Beijing Electronic Science and Technology Institute, China. His researches are focused on information security and visual media security.
\end{IEEEbiography}

\begin{IEEEbiography}[{\includegraphics[width=1in,height=1.25in,clip,keepaspectratio]{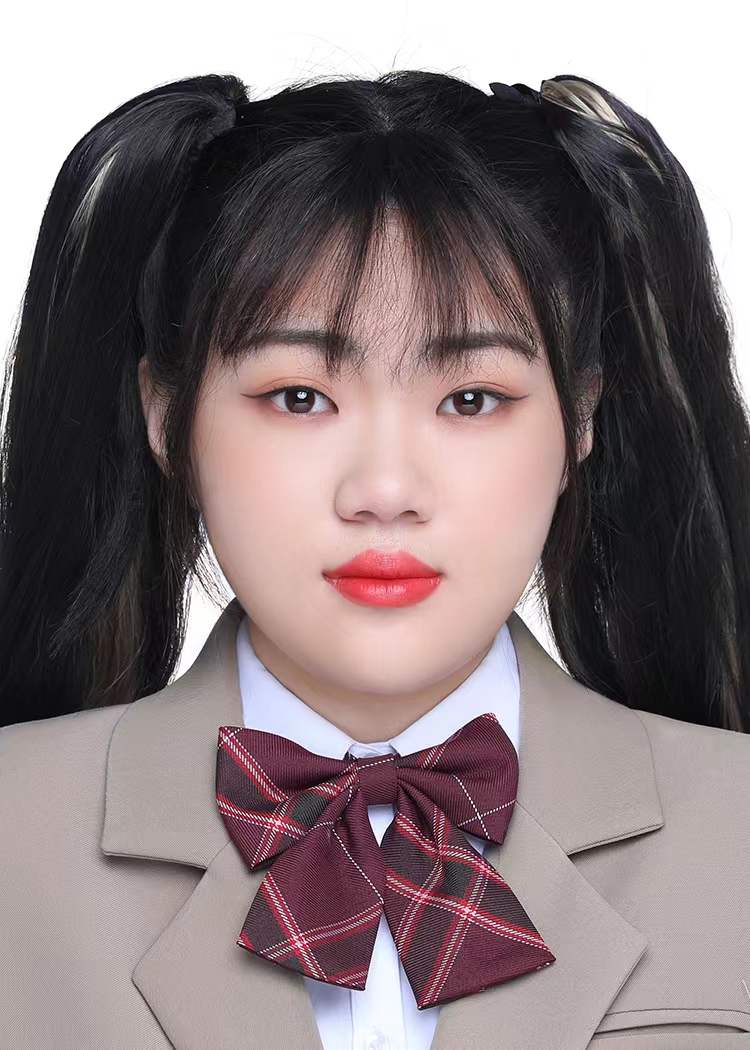}}]{SHUAI CUI}
	was born in Beijing, China. She received Bachelor of Arts and Sciences degree in philosophy and mathematics from University of California, Davis, CA, USA in 2021. Now, she is applying master programs. Her research interests include logic, metaphysics, epistemology, and Artificial Intelligence.
\end{IEEEbiography}

\begin{IEEEbiography}[{\includegraphics[width=1in,height=1.25in,clip,keepaspectratio]{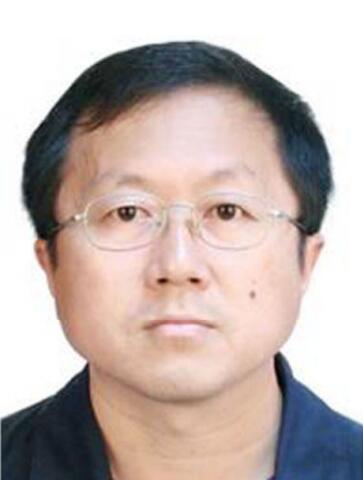}}]{XIAOKUN ZHANG}
	was born in Gansu, China. He is currently a Professor with the Department of Cyber Security, Beijing Electronic Science and Technology Institute, China. His research interest includes computer science and technology.
\end{IEEEbiography}

\begin{IEEEbiography}[{\includegraphics[width=1in,height=1.25in,clip,keepaspectratio]{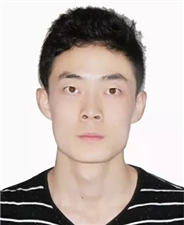}}]{XIQIAO LI}
	was born in Shanxi, China. He received the master’s degree from Beijing Electronic Science and Technology Institute, Beijing, China. His research interest includes visual media aesthetics.
\end{IEEEbiography}







\end{document}